\documentclass{article}

\usepackage{arxiv}

\usepackage[utf8]{inputenc} 
\usepackage[T1]{fontenc}    
\usepackage{hyperref}       
\usepackage{url}            
\usepackage{booktabs}       
\usepackage{amsfonts}       
\usepackage{nicefrac}       
\usepackage{microtype}      
\usepackage{graphicx}
\usepackage{amsmath}
\usepackage{caption}
\usepackage{subcaption}
\usepackage{algorithm}
\usepackage{algpseudocode}

\title{Sentiment analysis with genetically evolved Gaussian kernels}

\author{
Ibai Roman\\
Intelligent Systems Group\\
University of the Basque Country UPV/EHU\\
20018 Donostia, Spain\\
\texttt{ibai.roman@ehu.eus}\\
   \And
Alexander Mendiburu\\
Intelligent Systems Group\\
University of the Basque Country UPV/EHU\\
20018 Donostia, Spain\\
\texttt{alexander.mendiburu@ehu.eus}\\
   \And
Roberto Santana\\
Intelligent Systems Group\\
University of the Basque Country UPV/EHU\\
20018 Donostia, Spain\\
\texttt{roberto.santana@ehu.eus}\\
   \And
Jose A. Lozano\\
Intelligent Systems Group\\
University of the Basque Country UPV/EHU\\
20018 Donostia, Spain\\
Basque Center for Applied Mathematics BCAM\\
48009 Bilbao, Spain\\
\texttt{ja.lozano@ehu.eus}\\
}

\begin{document}
\maketitle

\begin{abstract}
Sentiment analysis consists of evaluating opinions or statements from the analysis of text. Among the methods used to estimate the degree in which a text expresses a given sentiment, are those based on Gaussian Processes. However, traditional Gaussian Processes methods use a predefined kernel with hyperparameters that can be tuned but  whose structure can not be adapted. In this paper, we propose the application of Genetic Programming for evolving Gaussian Process kernels that are more precise for sentiment analysis. We use use a very flexible representation of kernels combined with a multi-objective approach that simultaneously considers two quality metrics and the computational time spent by the kernels. Our results show that the algorithm can outperform Gaussian Processes with traditional kernels for some of the sentiment analysis tasks considered.
\end{abstract}


\section{Introduction} \label{sec:INTRO}

Among Natural Language Processing (NLP) problems, Sentiment Analysis (SA) has attracted much attention from the literature \cite{deriu_leveraging_2017,socher_semi-supervised_2011,beck_modelling_2017}. SA consists of understanding in an automated manner the opinion from written or spoken language. A common choice to solve these tasks is to compute a vector-based representation of the words, or embeddings, for all words in a sentence. Embeddings are then used by supervised learning algorithms to predict the presence or absence of an emotion in the text.

Kernel-based methods, due to their flexibility to capture different notions of similarity, have shown promising results inferring the latent sentiments from the embeddings \cite{beck_learning_2015,beck_learning_2017,beck_gaussian_2017}. Gaussian Process (GaussProc) models \cite{rasmussen_gaussian_2006} are one example of kernel-based methods which have been successfully applied to emotion or sentiment classification \cite{beck_modelling_2017}. GaussProcs rely on strong Bayesian inference foundations that allow them to update the model when new evidence is revealed. In comparison to other regression methods, they provide not only a prediction of a given function, but also an estimate uncertainty of the predictions. Apart from SA, GaussProcs also have been used in other NLP tasks, such as, text classification \cite{polajnar_protein_2011}, modeling periodic distributions of words over time \cite{preotiuc-pietro_temporal_2013}, and quality estimation \cite{cohn_modelling_2013,shah_investigation_2013}.

 
In all previous applications of GaussProc to NLP, the choice of the most appropriate kernel was made a-priori, among some well-known kernel functions. While there is a repertoire of kernels available in the literature \cite{rasmussen_gaussian_2006,duvenaud_automatic_2014,genton_classes_2002}, the selection of the most suitable one for a given problem is not straightforward. For example, for SA the most commonly applied kernel is the Squared Exponential (SE), also referred as the Radial Basis Function (RBF) kernel. However, Matern kernels have shown better results in this particular task \cite{beck_modelling_2017}.

Moreover, kernels usually have some parameters that need to be adjusted, which hardens the kernel selection problem. These parameters, often called hyperparameters, are usually tuned by maximizing a given metric (e.g., the marginal likelihood) \cite{blum_optimization_2013}.

Choosing the best metric to measure the quality of the kernels is not clear neither. While in the SA literature the Pearson's Correlation Coefficient (PCC) \cite{pearson_note_1895} has been used, this metric does not provide insights about the uncertainty that the GaussProc is able to model. As the GaussProcs offers a probabilistic prediction about the emotions, metrics such as the Log Marginal Likelihood (LML) \cite{rasmussen_gaussian_2006}, or the Negative Log Predictive Densities (NLPD) \cite{quinonero-candela_evaluating_2006}, seem better suited to choose between GaussProc models. 

In this paper we propose a method that does not rely on an a-priori specified kernel. This means that finding the kernel expression is part of the \emph{model selection} process. The learning algorithm that we use, which is based on Genetic Programming (GP) \cite{koza_genetic_1992}, is able to learn the kernel expression, together with an assignment of the corresponding hyperparameters. The evolved kernels keep the convenient property of being able to estimate the uncertainty of the predictions

Furthermore, by using a multi-objective approach the GP algorithm evaluates kernels using three different criteria. Two of these criteria measure kernel accuracy, and the third criterion minimizes the computational time needed by learning the kernel, therefore indirectly penalizing complexity. 

The remainder of the paper is organized as follows: The next section presents the addressed problem in the context of SA. Section~\ref{sec:gpr} introduces the main concepts related to GaussProc regression. The multi-objective GP approach to evolve kernel functions is presented in Section~\ref{sec:kernelsearch}. In Section~\ref{sec:relwork}, a review on related work is provided. We describe the experimental framework used to validate our algorithm, along with the numerical results in Section~\ref{sec:EXPE}. The conclusions of the paper and discussion of future work are presented in Section~\ref{sec:CONCLU}.

\section{Sentiment Analysis} \label{sec:sentiment}

Sentiment Analysis (SA) is an automated process that infers the opinion or feeling from a piece of text. It can be considered as particular type of semantic annotation of the text. SA is a very complex problem due to several factors including the ambiguity of human language, the large variability in the use of terms across individual, and the complexity of grammatical rules. However, the emergence of large text corpora and the usefulness of mining these corpora, e.g., for opinion mining related to products, services or politics  \cite{pang_opinion_2008}, have contributed to develop more advanced machine learning algorithms for this task. 

One of the directions of extending SA methods is to go beyond text categorization in positive or negative classes, to a more fine-grained emotion annotation \cite{strapparava_semeval-2007_2007}. This could be done by extending the number of classes in which a text is classified, but also by  allowing a continuous value of the strength in which the sentiment is manifested in the text. For  example, a subject is requested to evaluate, in a range $[0,100]$ a given sentiment (e.g., joy, fear, etc.) in a text. The problem we address in this paper is to automatically estimate this continuous value from the analysis of the text. This problem is posed as a supervised regression problem in which a number of annotated examples are available with the sentiment value.

\subsection{Sentence embeddings} \label{ssec:embeddings}

Machine learning methods require a representation of text in order to do semantic analysis.  There are a number of NLP approaches to represent and extract relevant information from text. In this domain, feature engineering to obtain informative features can be very labor-intensive. One increasingly used representation are word-vectors, or embeddings 

In a word embedding representation \cite{mikolov_efficient_2013}, each word is assigned a vector of continuous values. Embeddings are learned using neural network models. Some of the first embedding methods, and still among the most used ones, are the Continuous Bags of words (CBOW) and the Skip-gram models. The details on how these algorithms work is beyond the scope of this paper. 

Here, we focus on sentence embeddings, which in our case are computed as the mean of the embedding representations for all words in a given sentence. This is a common text representation for NLP tasks. For each sentence, we compute the embedding representation of all the words. Words missing in the dictionary are assigned a zero-vector representation. The average of the embedding representations of all the words  is computed to generate the sentence embedding. This sentence embedding is the representation from which we will estimate the sentiment value.

\section{Gaussian Process Regression} \label{sec:gpr}
A Gaussian Process (GaussProc) is a stochastic process, defined by a collection of random variables, any finite number of which have a joint Gaussian distribution \cite{rasmussen_gaussian_2006}. A GaussProc can be interpreted as a distribution over functions, and each sample of a GaussProc as a function.

GaussProcs can be completely defined by a mean function $m(\mathbf{x})$ and a covariance function, which depends on a kernel $k(\mathbf{x},\mathbf{x}')$. Given that, a GaussProc can be expressed as follows: 
\begin{equation} \label{eq:gp}
f(\mathbf{x}) \sim GaussProc(m(\mathbf{x}), k(\mathbf{x},\mathbf{x}'))
\end{equation}
where we assume that $\mathbf{x} \in \mathbb{R}^d$. We also consider an a-priori equal-to-zero mean function ($m(\mathbf{x})=0$), to focus on the kernel search problem.

A GaussProc can be used for regression by getting its posterior distribution given some (training) data. Thus, the GaussProc can provide a probabilistic model to infer the sentiment of an sentence embedding. The joint distribution between the training outputs $\mathbf{f}=(f_1,f_2,...,f_n)$ (where $f_i \in \mathbb{R}$, $i \in \{ 1,...,n \}$ and $n\in\mathbb{N}$) and the test outputs $\mathbf{f}_{*}=(f_{n+1},f_{n+2},...,f_{n+n_{*}})$ is given by:

\begin{equation} \label{eq:joint}
\begin{bmatrix}
\mathbf{f} \\ \mathbf{f}_{*}
\end{bmatrix} \sim \mathcal{N} \left( 0,
\begin{bmatrix}
K(X, X) & K(X, X_{*}) \\ K(X_{*}, X) & K(X_{*}, X_{*})
\end{bmatrix} \right)
\end{equation}
where $N(\mu,\Sigma)$ is a multivariate Gaussian distribution, $X=(\mathbf{x}_1,\mathbf{x}_2,...,\mathbf{x}_n)$ ($\mathbf{x}_i \in \mathbb{R}^d$, $i \in \{ 1,...,n \}$ and $n\in\mathbb{N}$) corresponds to the training inputs and $X_{*}=(\mathbf{x}_{n+1},  ..., \mathbf{x}_{n+n_{*}})$ to the test inputs. $K(X, X_{*} )$ denotes the $n \times n_{*}$ matrix of the covariances evaluated for all the $(X, X_{*})$ pairs.

The predictive Gaussian distribution can be found by obtaining the conditional distribution given the training data and the test inputs:
\begin{equation} \label{eq:gpr}
\begin{split}
\mathbf{f}_{*} | X_{*}, X, \mathbf{f} &\sim \mathcal{N}(\Hat{M}(X_{*}), \Hat{K}(X_{*}, X_{*}))\\
\Hat{M}(X_{*}) &= K(X_{*}, X)K(X, X)^{-1} \mathbf{f} \\
\Hat{K}(X_{*}, X_{*}) &= K(X_{*}, X_{*}) - K(X_{*}, X)K(X, X)^{-1}K(X, X_{*})
\end{split}
\end{equation}

\subsection{Covariance functions} \label{ssec:kernel}

GaussProc models use a kernel to define the covariance between any two function values \cite{duvenaud_automatic_2014}:

\begin{equation} \label{eq:kernel}
cov\left( f(\mathbf{x}), f(\mathbf{x}') \right) = k(\mathbf{x}, \mathbf{x}')
\end{equation}

The kernel functions used in GaussProcs are positive semi-definite (PSD) kernels. According to the Mercer's Theorem \cite{mercer_xvi._1909}, any PSD kernel can be represented as an inner product in some Hilbert Space.

The best known kernels in GaussProc literature are translation invariant, often referred to as stationary kernels. Among them, we focus on isotropic kernels where the covariance function depends on the norm:
\begin{equation} \label{eq:stationary}
k(\mathbf{x}, \mathbf{x}') = \hat{k}(r) \text{ where } r= \frac{1}{\theta_{l}} \left\lVert\mathbf{x}-\mathbf{x}'\right\rVert
\end{equation}
where $\theta_{l}$ is the lengthscale hyperparameter and $\hat{k}$ a function that guarantees that the kernel is PSD.

On the other hand, non-stationary kernels are the ones that may vary with translation. Within this family, the most common kernels are those that depend on the dot product of the input vectors. These kernels are usually referred to as dot-product kernels:
\begin{equation} \label{eq:dotproduct}
k(\mathbf{x}, \mathbf{x}') = \hat{k}(s) \text{  where  } s=\frac{1}{\theta_{l}}\left(\mathbf{x} - \theta_{s} \mathbf{1} \right) \left(\mathbf{x}' - \theta_{s} \mathbf{1} \right)^T
\end{equation}
where $\theta_{l}$ is again the lengthscale hyperparameter, $\theta_{s}$ is the shift hyperparameter and $\mathbf{1}$ is a vector of ones.

Table \ref{tab:kern} shows ten well-known kernels that have been previously used in the GaussProc literature.

\begin{table}[htbp]
 \centering
 \def\arraystretch{1.2}
 \begin{tabular}{|l|l|}
    \hline
    \multicolumn{2}{|c|}{Kernel function expressions} \\
    \hline
    Squared Exp. &  $\hat{k}_{SE} (r) = \theta_{0}^2 \ exp \left( -\frac{1}{2} r^2 \right)$  \\
    Matern $32$ &  $\hat{k}_{M32} (r) = \theta_{0}^2 \left( 1 + \sqrt{3} r \right) exp \left( - \sqrt{3} r \right)$  \\
    Matern $52$ &  $\hat{k}_{M52} (r) = \theta_{0}^2 \left( 1 + \sqrt{5} r + \frac{5}{3} r^2 \right) exp \left( -\sqrt{5} r \right) $  \\
    Rat. Quadratic &  $\hat{k}_{RQ} (r) = \theta_{0}^2 \left(1+ \frac{1}{2\alpha} r^2 \right)^{-\alpha}$ \\
    Exponential & $\hat{k}_{E} (r) = \theta_{0}^2 \ exp \left( - r\right) $\\
    $\gamma$-exponential &  $\hat{k}_{E\gamma} (r) = \theta_{0}^2 \ exp \left( - r^{\gamma} \right)$ \\
    Periodic &  $\hat{k}_{PER} (r) = \theta_{0}^2 \exp\left(-\frac{2\sin^2(\pi r)}{\theta_{p}^2}\right) $\\
    Linear &  $\hat{k}_{LIN} (s) = s $\\
    Constant & $k_{CON} (\mathbf{x}, \mathbf{x}') = \theta_{c} $\\
    White Noise & $k_{WN} (\mathbf{x}, \mathbf{x}') = \theta_{c} \ \delta(\mathbf{x}, \mathbf{x}') $\\
    \hline
  \end{tabular}
  \vspace{0.15cm}
  \caption{Well-known kernel functions. $\theta_{0}$ and $\theta_{p}$ are the kernel hyperparameters called amplitude and period respectively. $\delta$ is the Kronecker delta.}
  \label{tab:kern}
\end{table}

Some of these kernels have already been used also in NLP applications \cite{beck_learning_2017,beck_modelling_2017,specia_exploiting_2011}. The Squared Exponential (SE) kernel, described as $k_{SE}$ in the table, is known to capture the smoothness property of the objective functions. This kernel, was compared to Matern class kernels (denoted as $\hat{k}_{M32}$ and $\hat{k}_{M52}$ in the table) and the linear kernel ($\hat{k}_{LIN}$) in \cite{beck_modelling_2017}. There is little knowledge of the behaviour of other kernels when using GaussProcs in SA.


\subsection{Hyperparameter Optimization} \label{ssec:hpopt}

The choice of the kernel hyperparameters has a critical influence on the behavior of the model, and it is crucial to achieve good results in NLP applications of GaussProcs \cite{beck_modelling_2017}. This selection has been usually made by adjusting the hyperparameters of the kernel function so to optimize a given metric for the data. The most common approach is to find the hyperparameter set that maximizes the log marginal likelihood (LML):
\begin{equation} \label{eq:lml}
log \: p \left( \mathbf{f}| X, \boldsymbol\theta, K \right) = -\frac{1}{2} \mathbf{f}^T K(X,X)^{-1} \mathbf{f} - \frac{1}{2} log \: |K(X,X)| - \frac{n}{2} \: log \: 2\pi
\end{equation}
where $\boldsymbol\theta$ is the set of hyperparameters of the kernel and $n$ is the length of $X$.

\section{Evolving kernel functions} \label{sec:kernelsearch}

In this work we automatically search for new kernel functions in order to better predict sentiment expressed in a text. To guide this search, we propose to use a multi-objective GP approach. 

We encode the kernel functions by means of mathematical expression trees. Each expressions are defined in a strongly-typed grammar \cite{montana_strongly_1995} that specifies the possible combinations in which these kernels can be composed. We rely on the basic operations present in the well-known kernels of Table~\ref{tab:kern} (multiplication, square root, ...) to define the grammar.

These expression trees are evolved according to the GP approach shown in Algorithm~\ref{alg:moecov}. First, an initial population of $N$ kernels is generated. In order to do so, each individual is created at random. Each generation, the offspring is evaluated. Next, the relative improvement of each objective $O$ is measured. If the relative improvement in the current population is greater than a threshold $\beta$ for any of the objectives, a new population is generated through selection and variation.

After selecting the $\mu$ best individuals, the algorithm randomly chooses the variation method between a mutation or a crossover operator (with probability $p_{m}$ and $p_{cx}$ respectively, where $p_{cx}=1-p_{m}$) to generate an offspring population of $N$ new individuals. The next population is made up of the selected individuals and the offspring population. When the relative improvement is lower or equal to the threshold  $\beta$, the current population is replaced by a randomly generated one. This procedure is repeated for $G$ generations.  

\begin{algorithm}
  \begin{algorithmic}[1]
    \Procedure{MOECov}{$N$, $G$, $O$, $\mu$, $p_{m}$, $p_{cx}$, $\beta$}
      \State $offspring =$ \Call{GenRandPop}{$N$}
      \State $bestfit_{0} = (\infty)_{j=0}^{O}$
      \State $pop = offspring$
      \State $all = offspring$
      \State $i = 1$
      \While{$i < G$}
        \State \Call{Evaluate}{$offspring$}
      	\State $best =$ \Call{Select}{$pop$, $1$} 
        \State $bestfit_i = \Call{GetFitness}{best}$
        \State $relimprov = (\frac{bestfit_{i-1, j} - bestfit_{i,j}}{|bestfit_{i,j}|})_{j=0}^{O}$
        \If{$\beta < \Call{MAX}{relimprov}$} 
          \State $sel =$ \Call{Select}{$pop$, $\mu$}
          \State $offspring =$ \Call{Variate}{$sel$, $N$, $p_{m}$, $p_{cx}$}
          \State $bestfit_{i-1} = (\frac{1}{2} (bestfit_{i-1,j} + bestfit_{i,j})_{j=0}^{O}$
        \Else \Comment{Restart procedure}
          \State $sel = \emptyset$
          \State $offspring =$ \Call{GenRandPop}{$N$}
          \State $bestfit_{i-1} = (\infty)_{j=0}^{O}$
        \EndIf
        \State $pop = sel \cup offspring$
        \State $all = all \cup offspring$
        \State $i = i + 1$
      \EndWhile
      \State \Call{Evaluate}{$offspring$}
      \State $best =$ \Call{SelectBest}{$all$} 
      \State \Return $best$
    \EndProcedure
  \end{algorithmic}
  \caption{MOECov algorithm}
  \label{alg:moecov}
\end{algorithm}

\subsection{Random kernel creation} \label{ssec:randgen}
To randomly generate kernel expression trees that conform the initial population, we propose a grow method based on the work done in \cite{koza_genetic_1992}. The GENRANDPOP function in Algorithm~\ref{alg:moecov}, generates random trees by a recursive process where, at each step, a random terminal or operator is added in a type-safe manner.

\subsection{Variation operators} \label{ssec:variation}

Our kernel search method is based on perturbation or variation methods that modify previous solutions to obtain new ones. We use two variation operators, which are randomly selected every VARIATE function call in Algorithm~\ref{alg:moecov}: A crossover operator, which combines two kernel functions to generate a new one that keeps some of the features of its parents, and a mutation operator, which introduces slight modifications to the original kernel to obtain a new individual. We also explain how we control the depth of the trees generated by these variation methods.

We propose a crossover operator that randomly selects a subtree from each kernel and combines them with the sum or the product operator. This method merges the properties of each kernel while avoiding an excessive growth of the trees, known as bloating \cite{koza_genetic_1992}.

The mutation operator works by randomly selecting one of the following methods in a type-safe manner:
\begin{itemize}
  \item \textbf{Insert}: Inserts an elementary mathematical expression at a random position in the tree, as long as their output types agree. The subtree at the chosen position is used as the input of the created expression. If more inputs are required by the new primitive, new terminals are chosen at random.
  \item \textbf{Shrink}: This operator shrinks the expression tree by randomly choosing a branch and replacing it with one of the branch inputs (also randomly chosen) of the same type.
  \item \textbf{Uniform}: Randomly selects a point in the expression tree and replaces a subtree at that point by the subtree generated using our random generation method (Described in detail in Section~\ref{ssec:randgen}). Note that the output type of the random generated subtree must match the output type of the replaced one.
  \item \textbf{Node Replacement}: Replaces a randomly chosen operator from the kernel expression by an operator with the same number of inputs and types, also randomly chosen.
\end{itemize}

\subsection{Fitness evaluation} \label{ssec:fitness}

To estimate the quality of each kernel several metrics can be used. In SA tasks, Pearson correlation coefficient (PCC) is one of the most used metrics. It divides the covariance of two variables by the product of their standard deviations.
\begin{equation} \label{eq:PCC}
PCC(X, \mathbf{f})= \frac{\sum_{i=1}^{n}(f_i-\Hat{f})(\mu_i-\Hat{\mu})}{\sqrt{\sum_{i=1}^{n}(f_i-\Hat{f})^2\sum_{i=1}^{n}(\mu_i-\Hat{\mu})^2}}
\end{equation}
where $\Hat{f}$ is the mean of the vector $\mathbf{f}$ and $\Hat{\mu}$ is the mean of the posterior mean $\mathbf{\mu}$.

However, PCC does not take into account the probabilistic information given by the GaussProc.

On the other hand, the Negative Log Predictive Density (NLPD) \cite{quinonero-candela_evaluating_2006}, sums the likelihood of each prediction, and is more informative about the GaussProc's performance.
\begin{equation} \label{eq:NLPD}
NLPD(X, \mathbf{f}) = \frac{1}{n}\sum_{i=1}^{n}  -\frac{(f_i-\mu_i)^2}{2\sigma_i^2} - \frac{1}{2} log \: \sigma_i^2 - \frac{1}{2} \: log \: 2\pi
\end{equation}
where $\mu_i$ and $\sigma_i$ are the posterior mean and variance for $\mathbf{x}_{i}$.

Although both metrics are related, they evaluate the quality of the kernel from different perspectives. Ideally, it is desirable to guarantee a high correlation to the variable to be predicted along with a informative model of the uncertainty. Thus, the EVALUATE function in Algorithm~\ref{alg:moecov} measures both, PCC and NLPD metrics, being the objectives of our search. In addition, we also take the evaluation time as an third objective, to obtain kernels that are efficient in terms of the evaluation time.

\subsection{Multi-objective kernel selection} \label{ssec:selection}

We use a multi-objective selection operator based on the NSGA2 algorithm \cite{deb_fast_2000} to achieve good results in both metrics. As it can be seen in Algorithm~\ref{alg:moecov}, the SELECT function chooses $\mu$ individuals each generation. The population is divided into non-dominant groups, iteratively selecting the non-dominant group and repeating the operation with the rest of the individuals. The individuals in each group are sorted by crowding distance, and finally, the best $\mu$ individuals are chosen. To select the final best individual in the SELECTBEST function, the LML metric is used, in order balance the results of the objectives.




\subsection{Nested search procedure} \label{ssec:search}
\begin{figure}[htb]
  \begin{center}    
    \includegraphics[width=0.6\textwidth]{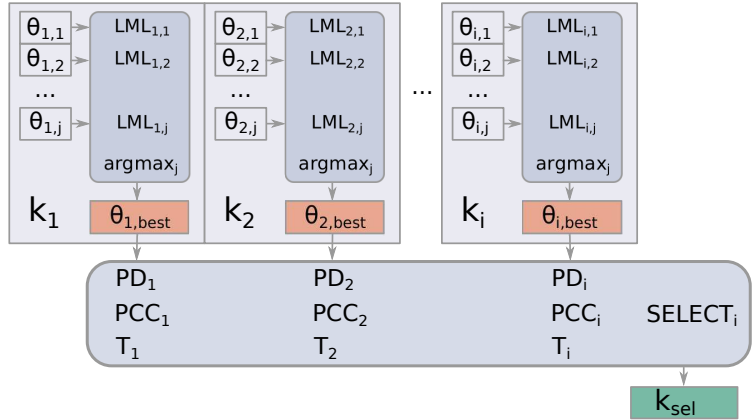}
    \caption{Two nested search procedures: The selection of the best hyperparameters for each kernel is made according to LML and the selection of the best kernel according to the BIC.}
    \label{fig:kernel_search}
  \end{center}    
\end{figure}

In contrast to other GP applications, the solutions in our approach do not encode all the necessary information to be evaluated. The optimal values of the hyperparameters, according to the LML, have to be determined. Thus, the performance of the solutions depends on the results of the hyperparameter optimization. Both search procedures, the selection of the best hyperparameters for each kernel and the selection of the best kernel given these hyperparameters, are illustrated in Figure~\ref{fig:kernel_search}.

In this paper, the hyperparameters are optimized by means of \textit{Powell}'s local search algorithm \cite{powell_efficient_1964}. As this algorithm is not bounded, the search space has to be constrained by penalizing non-feasible hyperparameter sets. On the other hand, as the function to optimize might be multi-modal, a multi-start approach was used, performing a random restart every time the stopping criteria of the \textit{Powell}'s algorithm are met, and getting the best overall result.  During this hyperparameter search, a maximum number of $150$ evaluations of the LML were allowed.
  
Once the hyperparameters were optimized, in order to obtain the fitness of each kernel, PCC and the NLPD were evaluated by dividing the training set into 3 cross-validation folds. The results in each fold were averaged to obtain the actual fitness of each kernel. Finally, the time spent measuring both metrics in each fold was summed as a third objective. 

Note that, as a result of the inclusion of the randomized restarts, the hyperparameters found for a certain kernel in two independent evaluations may not be the same. In fact, this implies that the fitness function optimized by the GP algorithm is stochastic.

\section{Related work} \label{sec:relwork}

Word embeddings \cite{mikolov_distributed_2013} are extensively applied to NLP tasks \cite{lampos_enhancing_2017}. The usual approach  when combining embeddings from words in a sentence is to compute the average. This is the procedure used in \cite{beck_modelling_2017}, where 100-dimensional Glove embeddings \cite{pennington_glove:_2014} are the representation of choice for mapping texts to emotion scores using GaussProc regression.

GaussProcs are particularly suited to model uncertainty in the predictions and allow to accurately predict under noisy conditions. As such, there are diverse scenarios in which GaussProc can be applied to NLP tasks \cite{cohn_gaussian_2014}. In all these problems the kernel function was selected beforehand, although its choice varies depending on the problem. The most frequently used kernel is the RBF kernel. However, this kernel is not suitable for all the problems. In \cite{preotiuc-pietro_temporal_2013}, text periodicities of tweets are modeled using GaussProc with kernels specifically suited  to capture the periodicities of the tweets. In the same paper, the periodic kernel (PER) and the periodic spike (PS) are shown to outperform non-periodic kernels and capture different periodic patterns. In \cite{beck_modelling_2017}, where GaussProcs are applied to SA, four different kernels are compared: SE, Linear, two Matern kernels. The Matern kernels are reported to produce better results than SE. In addition to numerical kernels, structural-kernels (e.g., tree-kernels)  have been also combined with GaussProc.


Research on the evolution of kernel functions using evolutionary algorithms have shown that it is also possible to explore the space of kernel functions beyond the hyperparameter optimization. In the GaussProc literature this has been done by combining known kernels \cite{kronberger_evolution_2013,duvenaud_structure_2013,lloyd_automatic_2014}. Kernels have also been evolved for Support Vector Machines (SVMs) \cite{howley_evolutionary_2006,gagne_genetic_2006,diosan_evolving_2007, sullivan_evolving_2007,diosan_improving_2012,koch_tuning_2012} and Relevance Vector Machines (RVMs) \cite{bing_gp-based_2010}. Some of the SVM approaches are also based in combining the well-known kernels \cite{sullivan_evolving_2007,diosan_improving_2012}, although in some other works the kernels are learned from simple mathematical expressions \cite{howley_evolutionary_2006,gagne_genetic_2006,diosan_evolving_2007,koch_tuning_2012}.

The particular characteristics determined by the GaussProcs make the evolution of kernels in this domain different to those algorithms used for SVMs and RVMs. Our approach considers different ways to evaluate the kernels based on a multi-objective approach, and creates kernels for GaussProc from scratch, without seeding components of human-designed kernels. This characteristic allows us to derive kernels that are not constrained by the prior knowledge while at the same time being optimized for the desired objectives.

\section{Experiments} \label{sec:EXPE}

The goal of our experiments is to evaluate the performance of the introduced algorithm for the task of sentiment prediction from text.  First we introduce the problem benchmark and word embeddings to evaluate the algorithm. Then, we describe the parameters used by the algorithms and explain the characteristics of the experimental framework, including the metrics used to compare the algorithms. Finally, we present the numerical results obtained from the experiments and discuss these results. 

\subsection{Problem benchmark and word-embeddings} \label{sec:datasets}

We use the SemEval2007 Affective Text shared task dataset \cite{strapparava_semeval-2007_2007}\footnote{Available at \url{https://web.eecs.umich.edu/\~mihalcea/downloads.html\#affective}}, following the work done in \cite{beck_modelling_2017}. In this dataset, news headlines were manually annotated by experts, assigning to each text a degree of presence for each six Eckman~\cite{ekman_facial_1993} emotions: anger, disgust, fear, joy, sadness and surprise. In the original work, where this dataset was introduced, texts were divided into  "dev" and "test" datasets. For the experiments presented in  \cite{beck_modelling_2017}, the two sets were combined and further divided into $10$ folds used for cross-validation. We use this 10-fold partition to evaluate our algorithms. 

In order to compute a representation for each text, punctuations in each headline were removed, tokenized \cite{bird_natural_2009}, and case ignored.  From the resulted text, the word-vector representation of each word were obtained using the 100-dimensional GloVe embeddings \cite{pennington_glove:_2014}\footnote{Available from \url{https://nlp.stanford.edu/projects/glove/}}.  After deleting the words that were not found in the embedding,  the representation of each headline was computed as the average of the words.

\subsection{Experimental setup} \label{sec:expsetup}

Our experiments consists of learning a kernel for a GP regressor that predicts a particular emotion based on sentence embeddings. We evaluate the quality of the final kernels in terms of the PCC and  NLPD metrics.  

We compare MOECov algorithm with different variants of a-priori defined kernel methods, which were shown in  \cite{beck_modelling_2017} to produce good prediction results for the six emotions previously described. The kernels are presented in Table~\ref{tab:kern}. 

 The parameters used by the evolutionary algorithm were:
\begin{itemize}
    \item Population size: $N=38$,
    \item Number of generations: $G=65$,
    \item  $O=2$,
    \item  Mutation ($p_{m}=0.4$) and crossover ($p_{cx}=0.6$) probabilities,
    \item  $\mu=9$,
    \item  $\beta=1\mathrm{e}{-5}$ 
\end{itemize}

All algorithms were coded in Python.  The implementation of MOECov\footnote{This implementation is freely available from \emph{Omitted for blind review}} is based on the EA software \texttt{DEAP}\footnote{\url{https://deap.readthedocs.io}} \cite{fortin_deap:_2012}.

Due to the stochastic nature of the algorithm, for all algorithms and in every dataset, the kernel search process was repeated $30$ times along $10$ random cross-validation folds.


\subsection{Results of the comparison between the algorithms} \label{sec:results}


Table~\ref{tab:PCC} shows the average PCC metric of the best solution obtained by the algorithms in the $30$ experiments. We remark that, these values have been computed on the test data, which we have not used for learning the GP programs. The results, for the NLPD metrics are shown in Table~\ref{tab:NLPD}. In the tables, the best average value obtained for each of the sentiments are highlighted. 

The analysis of the tables reveals that, in terms of the average fitness,  MOECov improves all other kernels for both metrics in most of the cases. For the PCC metric, only in the \texttt{disgust} dataset, MOECov was not able to outperform the M52 kernel. For the NLPD metric, MOECov outperforms all the algorithms for all the sentiments. 
 
Among the well-known kernels, M52 seems a better choice than the others, as it gets the second best result in average in the rest of the problems according to the PCC metric. M52 is also the second best choice in NLPD, only surpassed by SE kernel in the \texttt{fear} dataset, while the results of  M32 and SE kernels are similar. As it can be appreciated in the tables, the LIN kernel is the worst performing kernel according to the average results.



\begin{table}[hbtp]
\centering
\begin{tabular}{|l|r|r|r|r|r|}
\hline
\multicolumn{1}{|c}{} & \multicolumn{1}{|c}{LIN} & \multicolumn{1}{|c}{M32} & \multicolumn{1}{|c}{M52} & \multicolumn{1}{|c}{SE} & \multicolumn{1}{|c|}{MOECov} \\
\hline
\hline
anger & 0.58592 & 0.63556 & 0.6401 & 0.62629 & \textbf{0.64690}\\
disgust & 0.44828 & 0.52492 & \textbf{0.52782} & 0.50111 & 0.52456\\
fear & 0.68056 & 0.728096 & 0.73059 & 0.72737 & \textbf{0.73555}\\
joy & 0.53832 & 0.55775 & 0.57459 & 0.56341 & \textbf{0.59158}\\
sadness & 0.63625 & 0.67148 & 0.68205 & 0.67876 & \textbf{0.69710}\\
surprise & 0.40311 & 0.45416 & 0.45647 & 0.43758 & \textbf{0.46751}\\
\hline
\end{tabular}
\caption{Mean results for PCC metric. Best results are shown in bold.}
\label{tab:PCC}
\end{table}

\begin{table}[hbtp]
\centering
\begin{tabular}{|l|r|r|r|r|r|}
\hline
\multicolumn{1}{|c}{} & \multicolumn{1}{|c}{LIN} & \multicolumn{1}{|c}{M32} & \multicolumn{1}{|c}{M52}  & \multicolumn{1}{|c}{SE} & \multicolumn{1}{|c|}{MOECov} \\
\hline
\hline
anger & 3.94141 & 3.92037 & 3.91162 & 3.93041 & \textbf{3.91084}\\
disgust & 3.81476 & 3.78148 & 3.77491 & 3.80068 & \textbf{3.77419}\\
fear & 4.16636 & 4.10615 & 4.1006 & 4.0986 & \textbf{4.07623}\\
joy & 4.34633 & 4.32588 & 4.30362 & 4.32737 & \textbf{4.29549}\\
sadness & 4.31082 & 4.28845 & 4.27618 & 4.28176 & \textbf{4.24454}\\
surprise & 4.06292 & 4.04524 & 4.04519 & 4.0511 & \textbf{4.02712}\\
\hline
\end{tabular}
\caption{Mean results for NLPD metric. Best results are shown in bold.}
\label{tab:NLPD}
\end{table}

We conducted a statistical test to assess the existence of significant differences among the algorithms. For each metric and emotion, we applied the Friedman's test \cite{friedman_use_1937} and we found significant differences in every comparison (p-values can be seen in the figures). Then, for each configuration, we applied a post-hoc test based on Friedman’s test, and adjusted its results with the Shaffer's correction \cite{shaffer_modified_2012}.

\begin{figure}[htb]
  \begin{center}
    \begin{subfigure}[t]{0.28\textwidth}
        \centering
        \includegraphics[width=\linewidth,trim={70 70 50 70},clip]{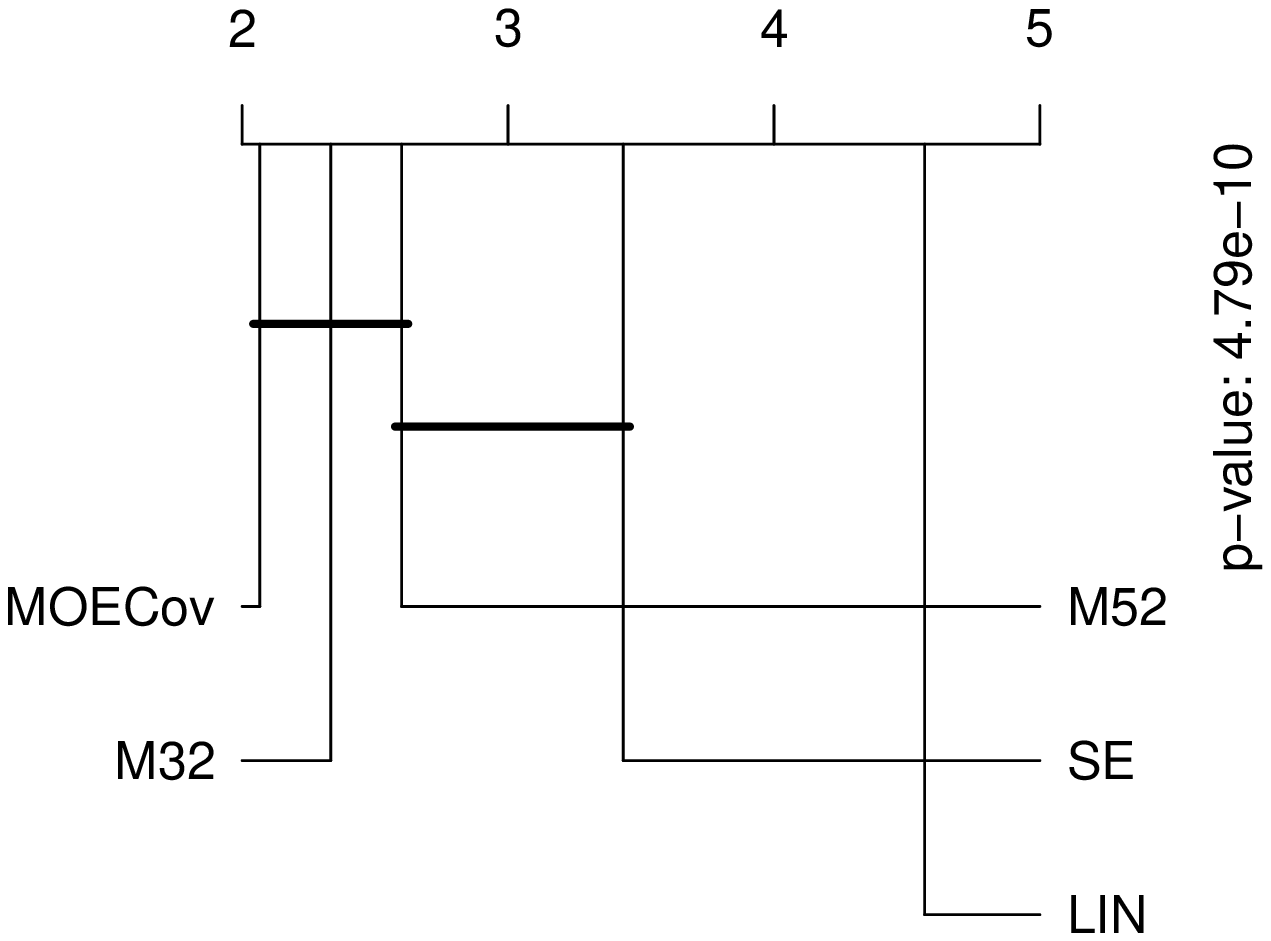}
        \caption{Anger}
        \label{fig:pcccritdiffanger}
    \end{subfigure}%
    \begin{subfigure}[t]{0.28\textwidth}
        \centering
        \includegraphics[width=\linewidth,trim={70 70 50 70},clip]{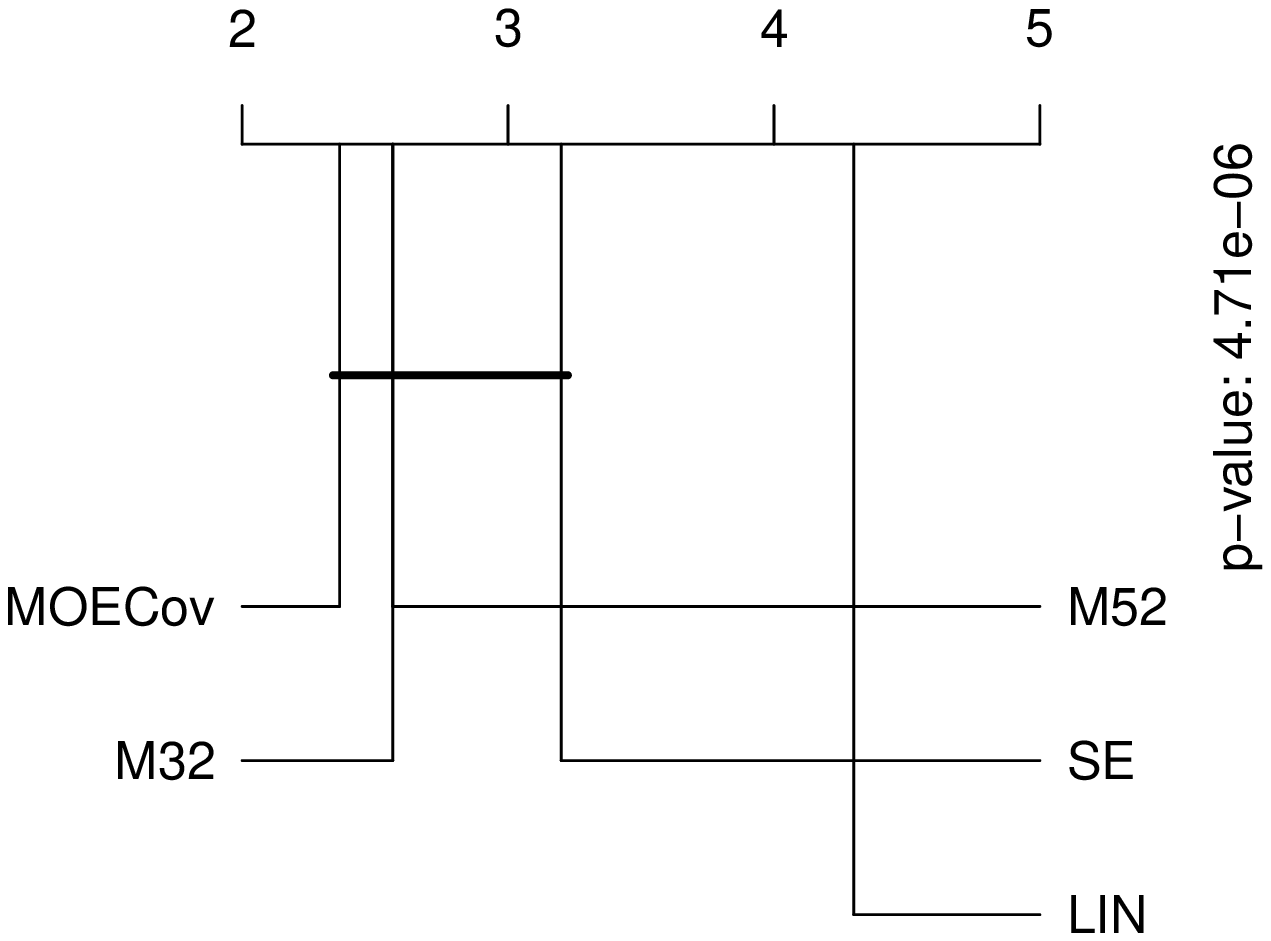}
        \caption{Disgust}
        \label{fig:pcccritdiffdisgust}
    \end{subfigure}%
    \begin{subfigure}[t]{0.28\textwidth}
        \centering
        \includegraphics[width=\linewidth,trim={70 70 50 70},clip]{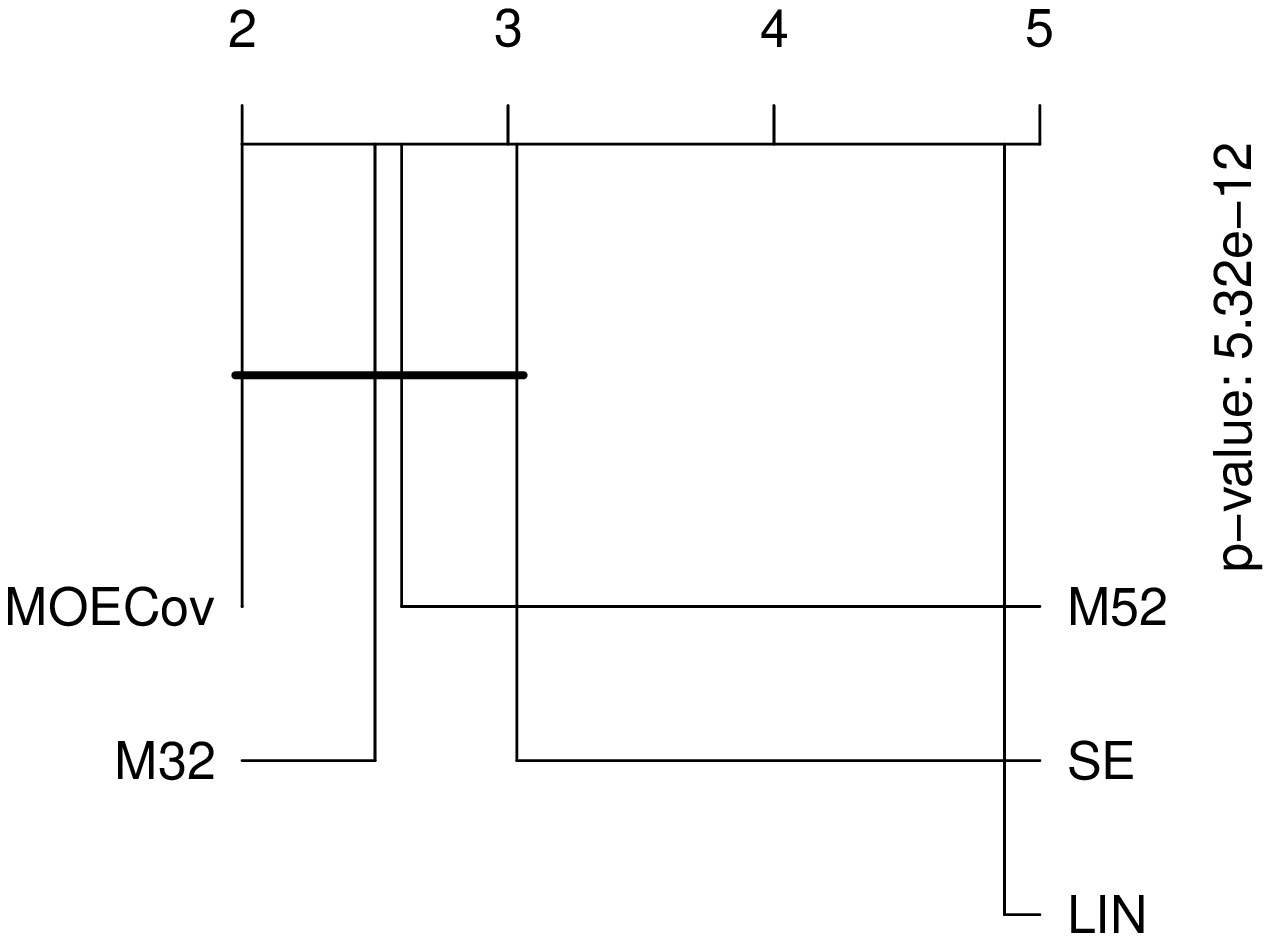}
        \caption{Fear}
        \label{fig:pcccritdifffear}
    \end{subfigure}\\
    \begin{subfigure}[t]{0.28\textwidth}
        \centering
        \includegraphics[width=\linewidth,trim={70 70 50 70},clip]{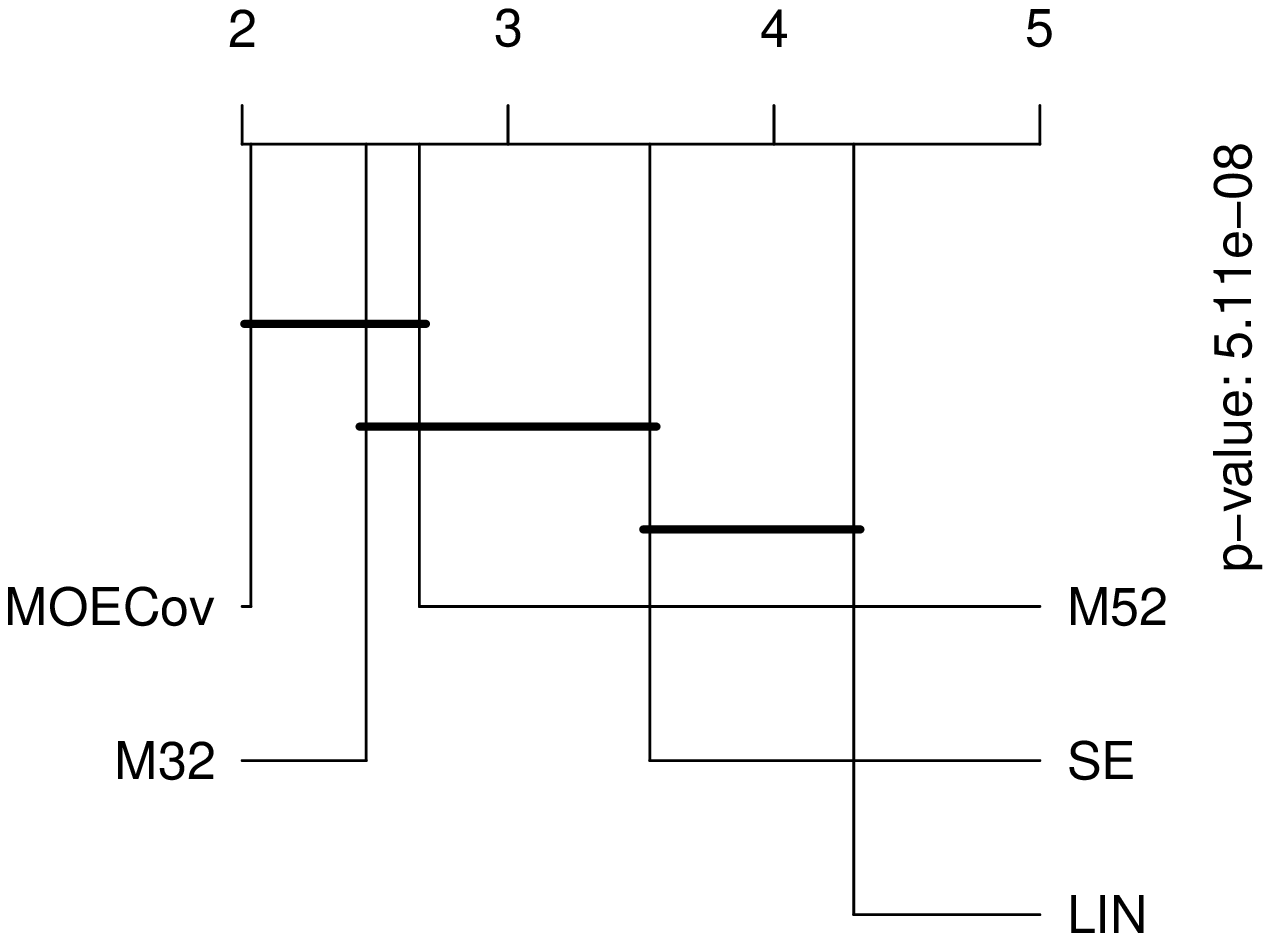}
        \caption{Joy}
        \label{fig:pcccritdiffjoy}
    \end{subfigure}%
    \begin{subfigure}[t]{0.28\textwidth}
        \centering
        \includegraphics[width=\linewidth,trim={70 70 50 70},clip]{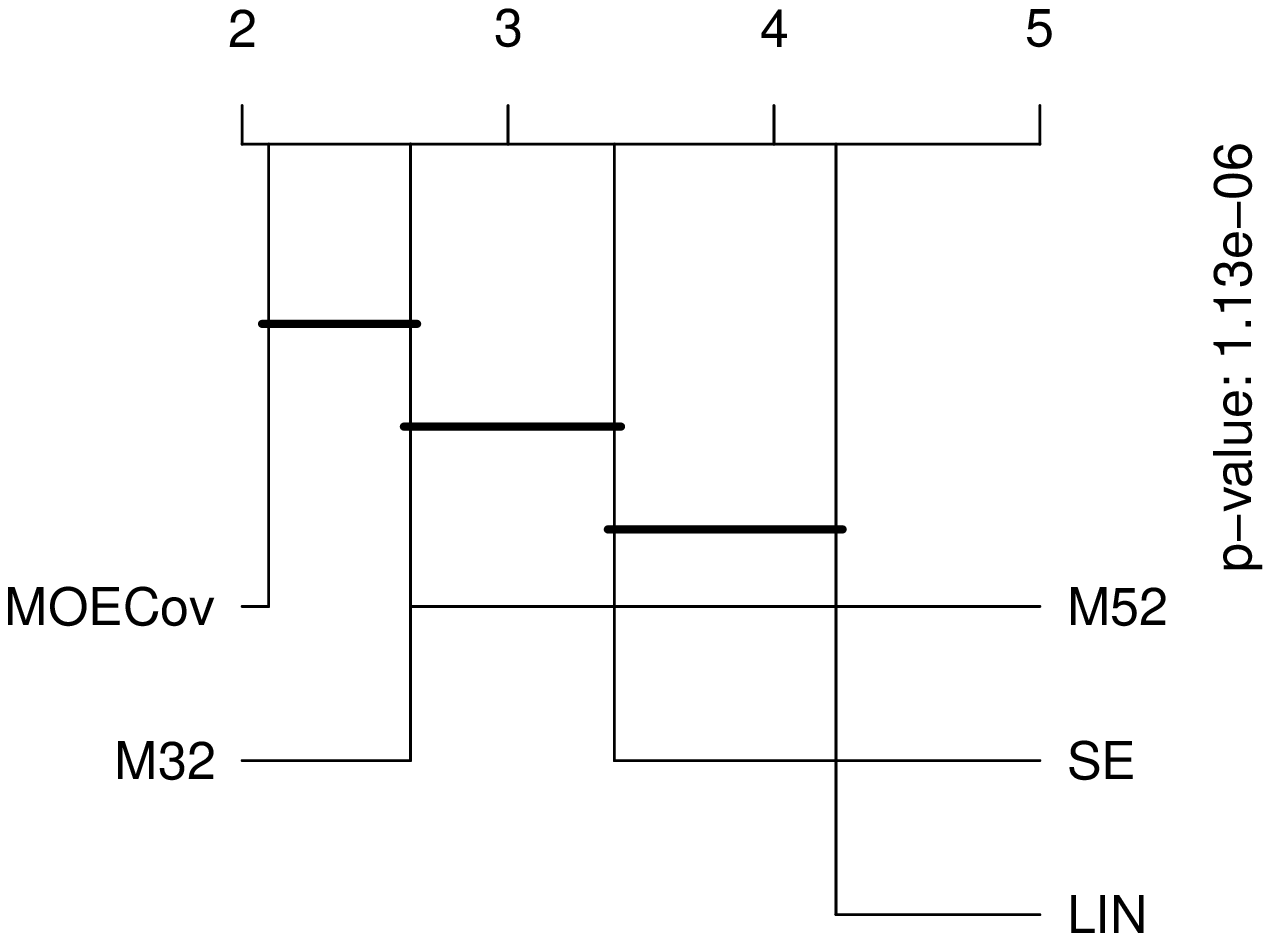}
        \caption{Sadness}
        \label{fig:pcccritdiffsadness}
    \end{subfigure}%
    \begin{subfigure}[t]{0.28\textwidth}
        \centering
        \includegraphics[width=\linewidth,trim={70 70 50 70},clip]{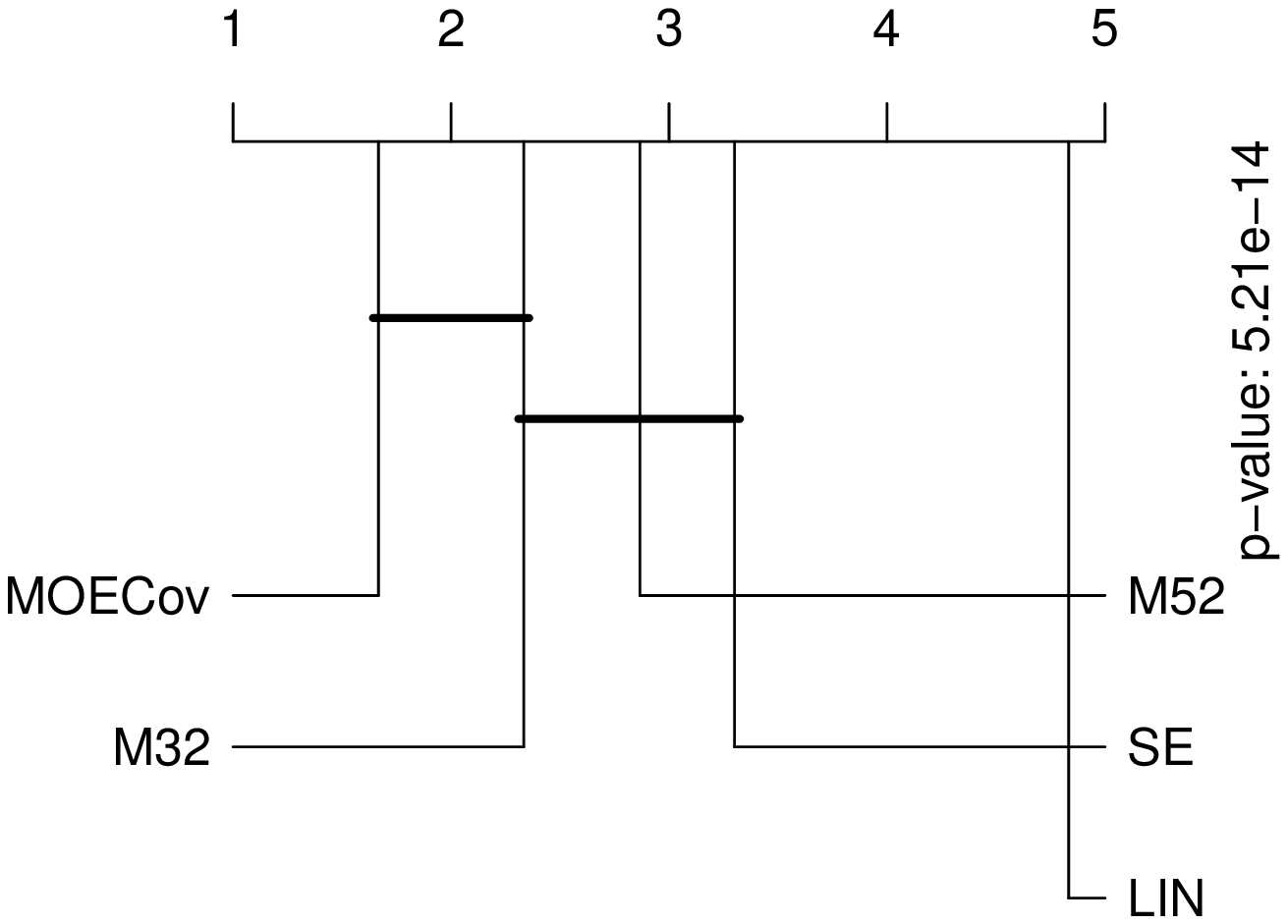}
        \caption{Surprise}
        \label{fig:pcccritdiffsurprise}
    \end{subfigure}\\
    \caption{Critical difference diagrams for the PCC metric. The kernels are ordered following the results in their ranking. The metrics with no significant differences between them are matched with a straight line.}
    \label{fig:pcccritdiff}
  \end{center}    
\end{figure}

\begin{figure}[htb]
  \begin{center}
    \begin{subfigure}[t]{0.28\textwidth}
        \centering
        \includegraphics[width=\linewidth,trim={70 70 50 70},clip]{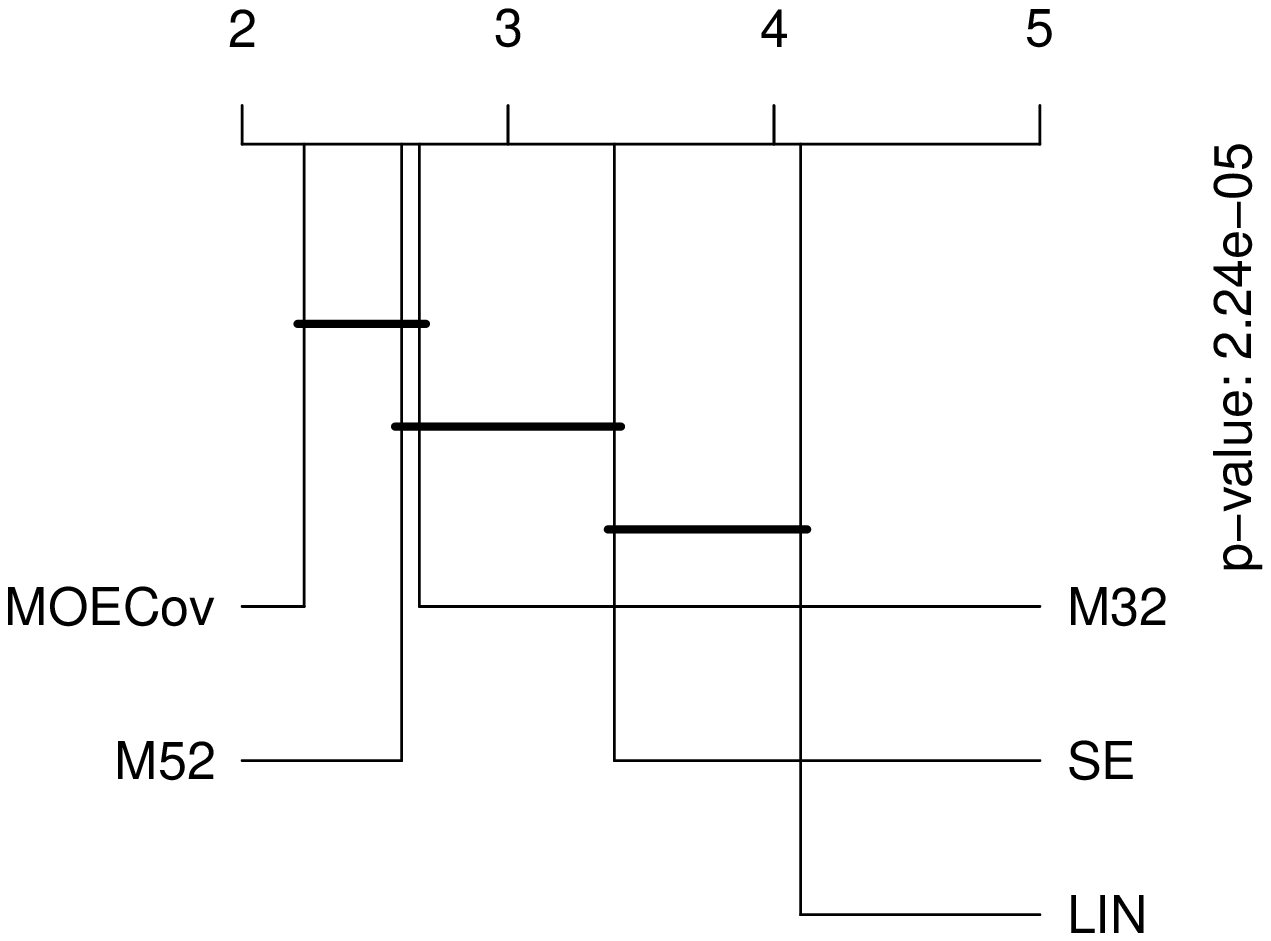}
        \caption{Anger}
        \label{fig:nlpdcritdiffanger}
    \end{subfigure}%
    \begin{subfigure}[t]{0.28\textwidth}
        \centering
        \includegraphics[width=\linewidth,trim={70 70 50 70},clip]{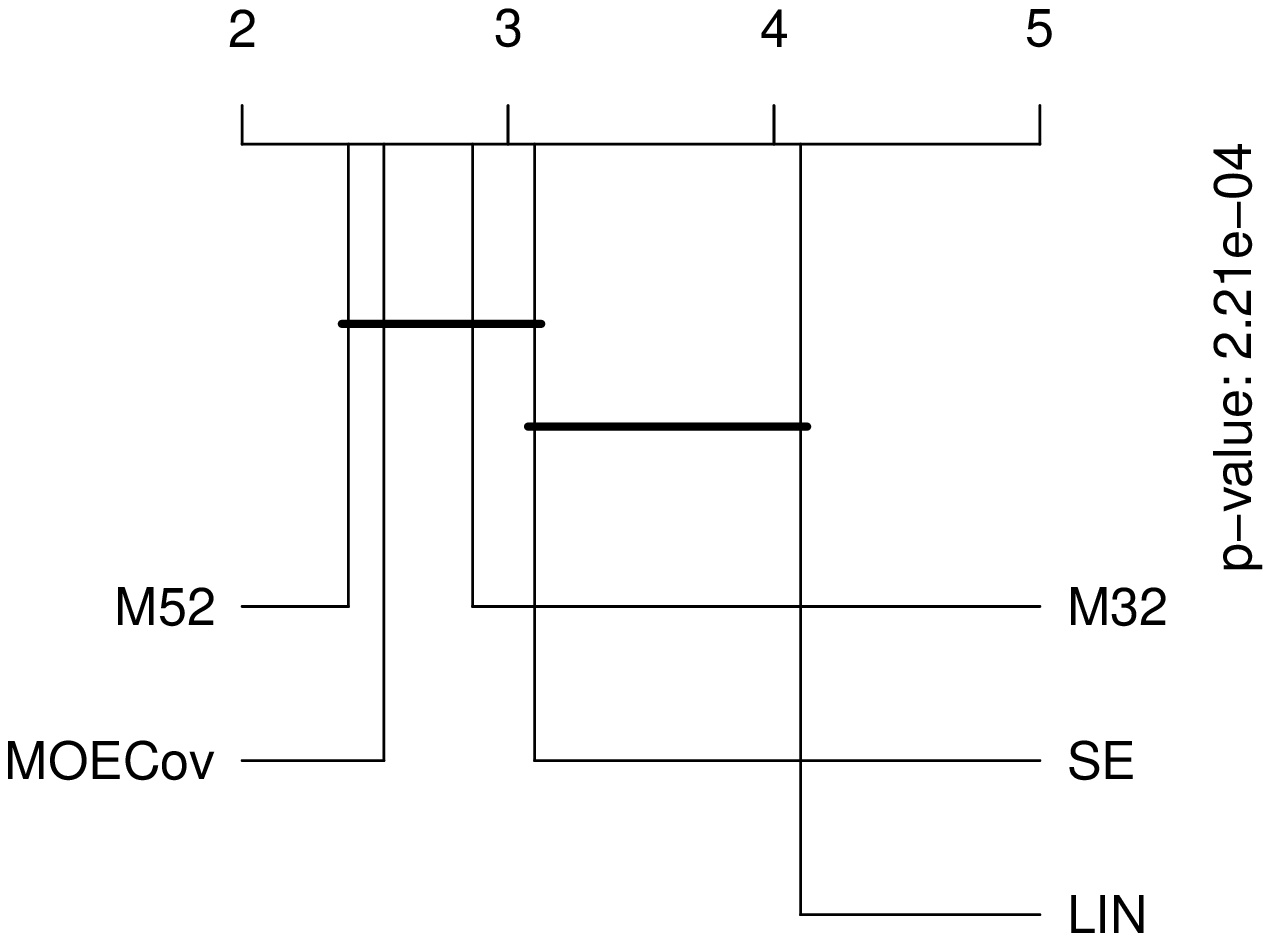}
        \caption{Disgust}
        \label{fig:nlpdcritdiffdisgust}
    \end{subfigure}%
    \begin{subfigure}[t]{0.28\textwidth}
        \centering
        \includegraphics[width=\linewidth,trim={70 70 50 70},clip]{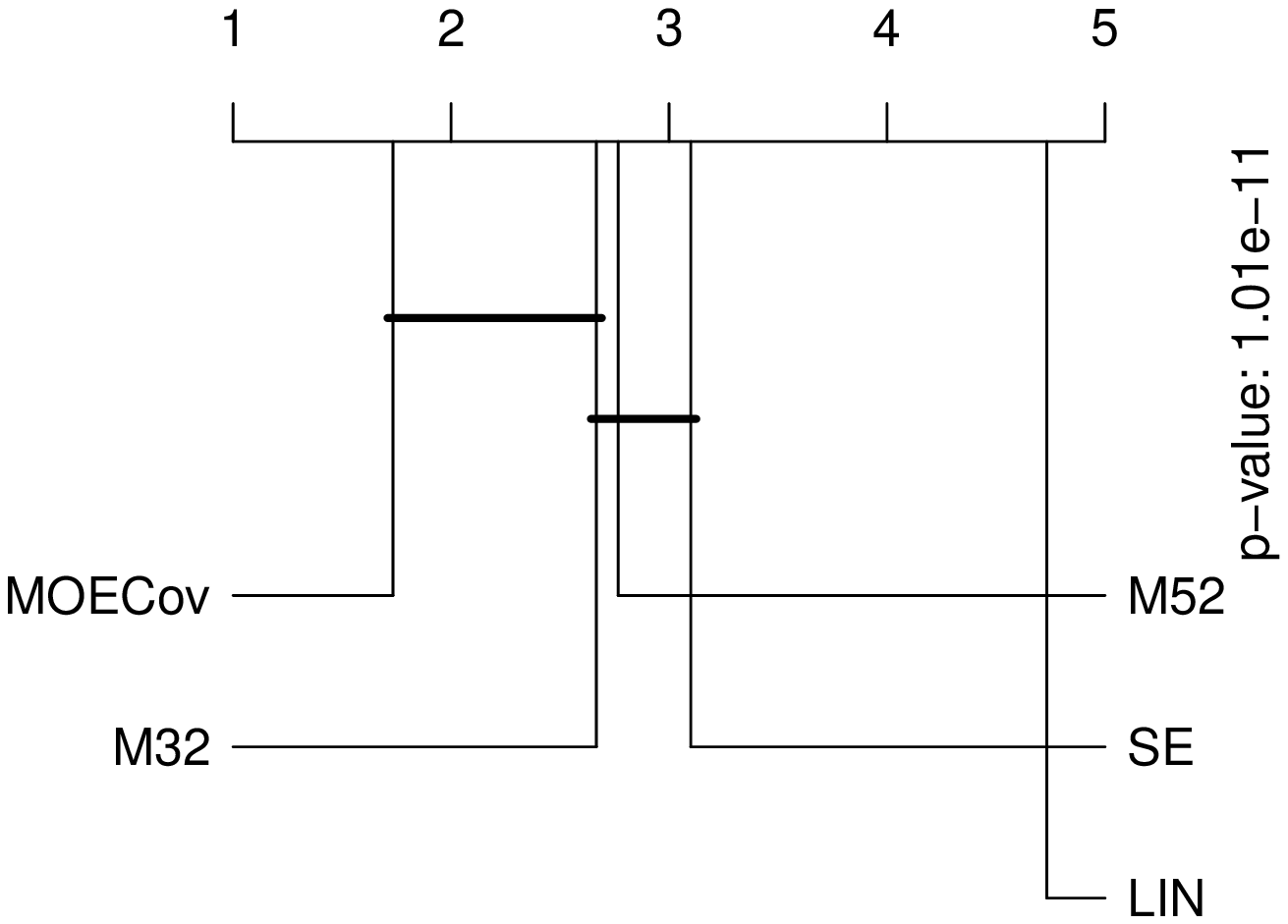}
        \caption{Fear}
        \label{fig:nlpdcritdifffear}
    \end{subfigure}\\
    \begin{subfigure}[t]{0.28\textwidth}
        \centering
        \includegraphics[width=\linewidth,trim={70 70 50 70},clip]{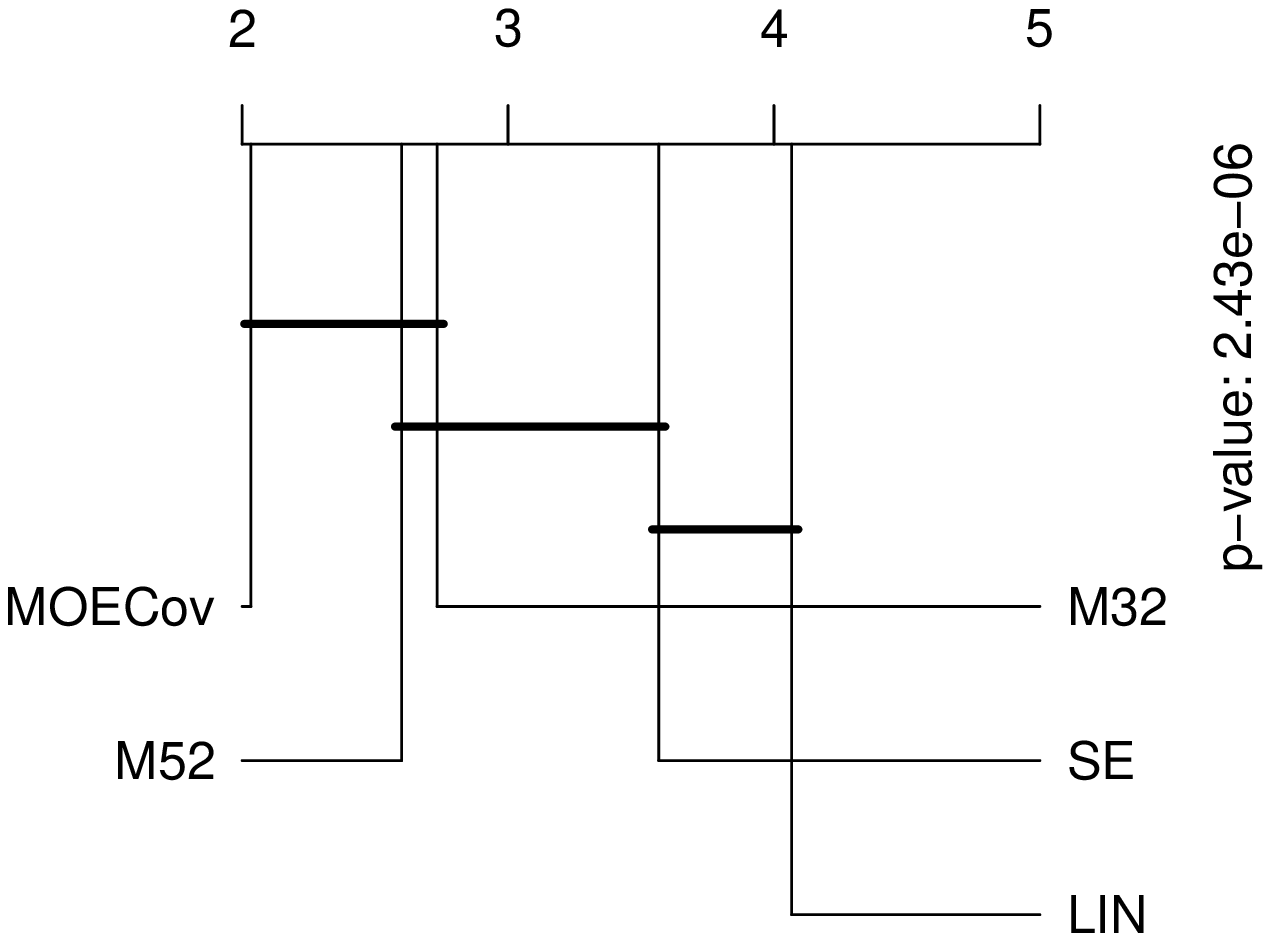}
        \caption{Joy}
        \label{fig:nlpdcritdiffjoy}
    \end{subfigure}%
    \begin{subfigure}[t]{0.28\textwidth}
        \centering
        \includegraphics[width=\linewidth,trim={70 70 50 70},clip]{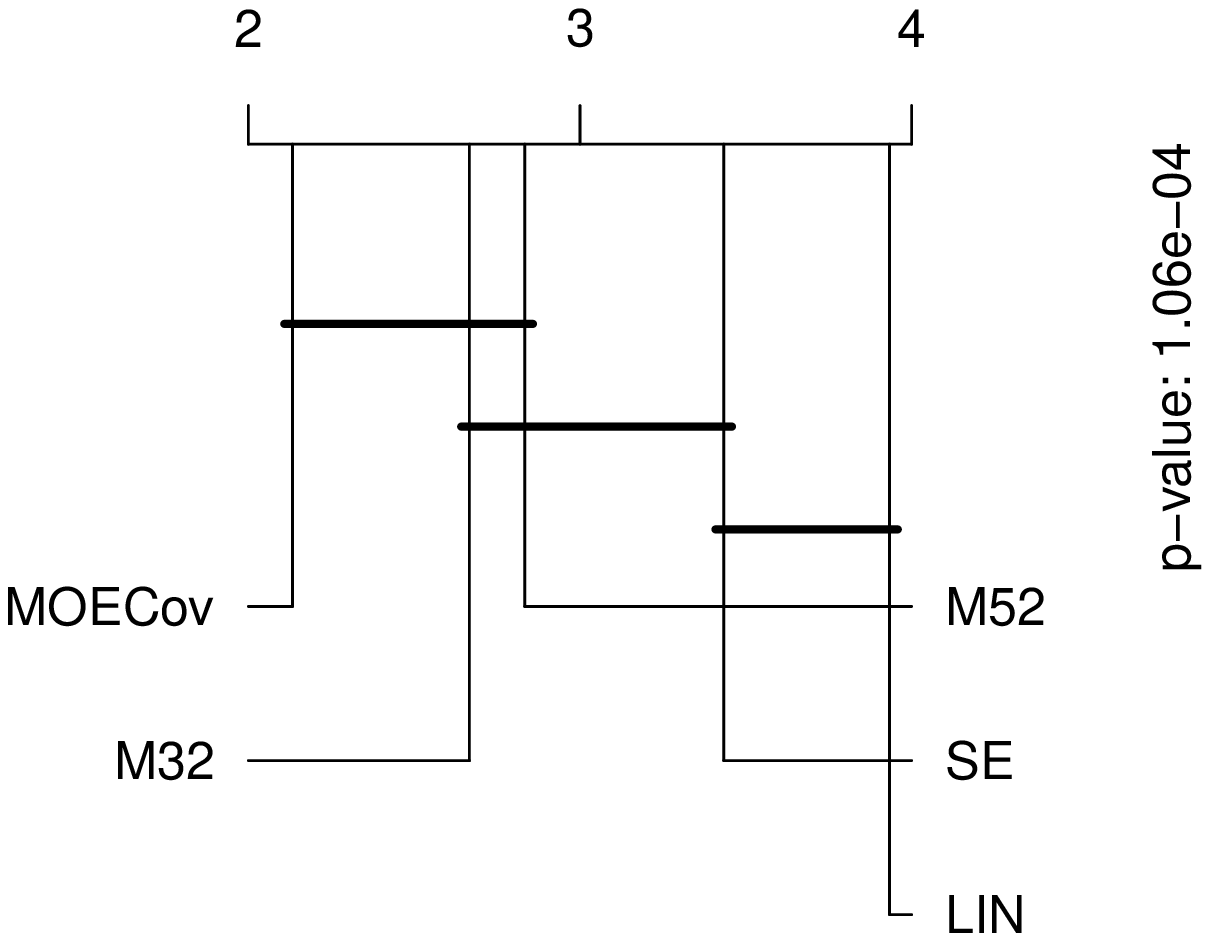}
        \caption{Sadness}
        \label{fig:nlpdcritdiffsadness}
    \end{subfigure}%
    \begin{subfigure}[t]{0.28\textwidth}
        \centering
        \includegraphics[width=\linewidth,trim={70 70 50 70},clip]{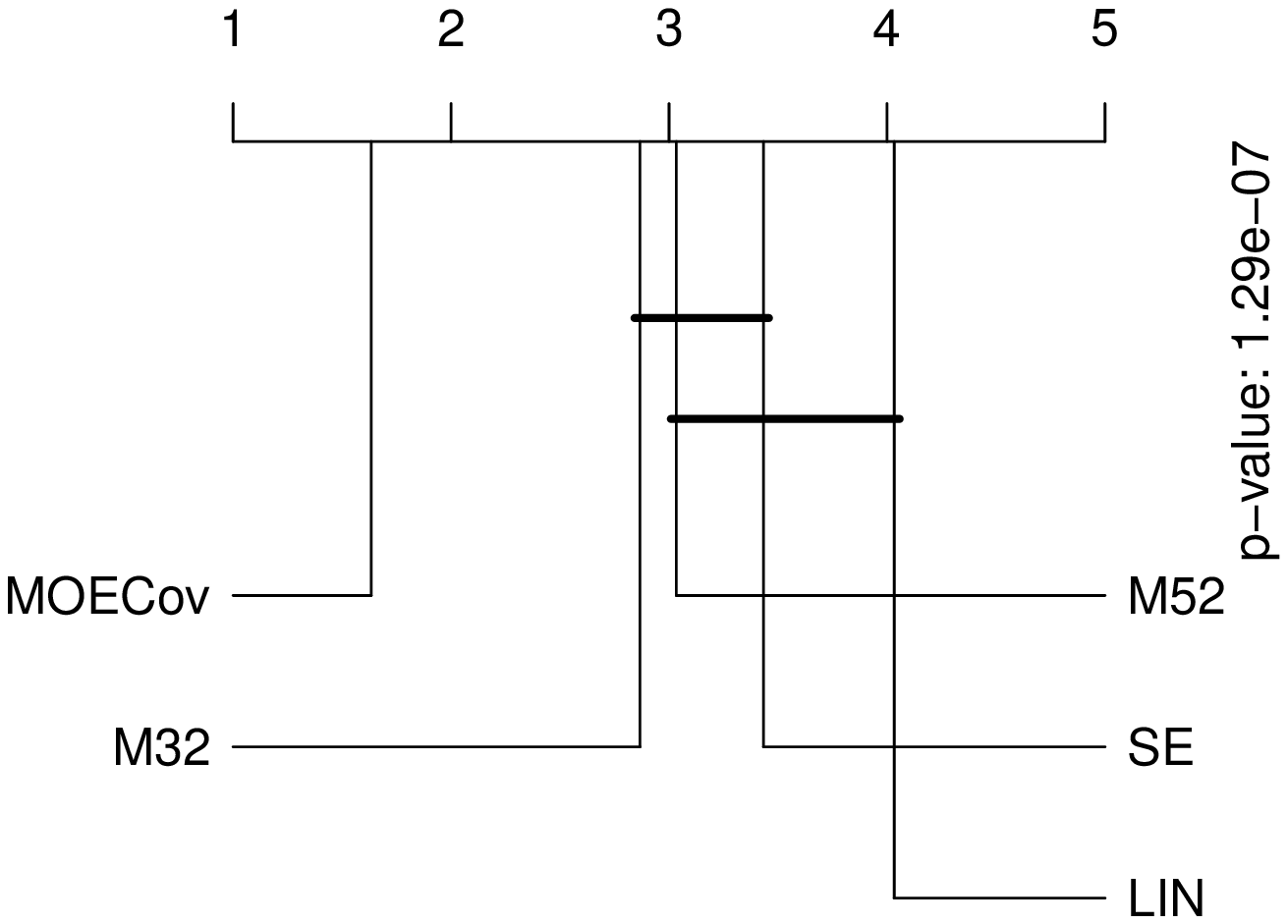}
        \caption{Surprise}
        \label{fig:nlpdcritdiffsurprise}
    \end{subfigure}\\
    \caption{Critical difference diagrams for the NLPD metric. The kernel are ordered following the results in their ranking. The metrics with no significant differences between them are matched with a straight line.}
    \label{fig:nlpdcritdiff}
  \end{center}    
\end{figure}

The results are shown in Figure~\ref{fig:pcccritdiff} and Figure~\ref{fig:nlpdcritdiff}. The results confirm a coherent pattern where MOECov is the best performing algorithm for most of the datasets. However, it is also appreciated that according to this test, for most of the datasets, the differences between MOECov and M52 are not significant. In evaluating these results it is important to take into account that the kernels produced by MOECov have been generated completely from scratch, with no prior knowledge of the existing kernels. The algorithm is able evolve a well performing kernel starting from elementary mathematical components.

\subsection{Analysis of the MOECov evolution} \label{sec:MOECov_results}

\begin{figure}[htb]
  \begin{center}
    \includegraphics[width=0.92\linewidth]{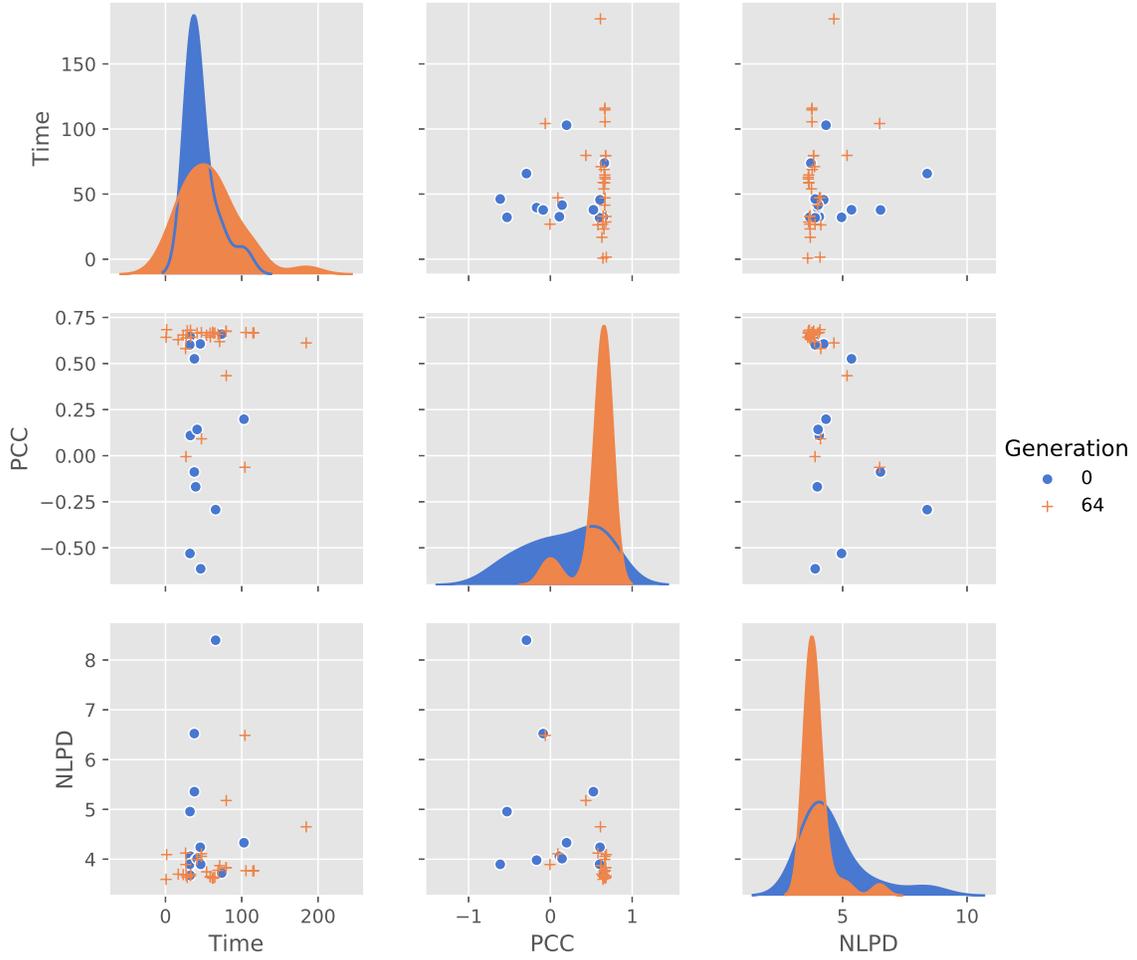}
    \caption{Distribution of the objective values in the first and last population of MOECov for one execution of the algorithm in the \texttt{anger} dataset.}
    \label{fig:MOECov_First}
  \end{center}    
\end{figure}

One characteristic feature of our approach is that we simultaneously optimize different characteristics of the kernels. In order to determine whether the multi-objective approach effectively leads to the creation of more efficient kernels, both in terms of the accuracy for the prediction task, and in terms of efficiency, we analyze the fitness distribution of the solutions in the first and last population of MOECov for the \texttt{anger} dataset. These results are shown in Figure~\ref{fig:MOECov_First} where, in order to ease the visualization, results are shown for only one execution of the algorithm. 
 
In Figure~\ref{fig:MOECov_First} we represent the scatter plots for each possible pair of objectives. As it can be appreciated in the figures, from the first to the last population there is an improvement in the values of the objective values for the PCC and NLPD metrics. However, the computational time actually increases from the first to the last population. This result is not surprising since it is expected that the increase in accuracy of the trees is achieved by also augmenting their complexity. In this scenario we expect that using the time as a third objective can serve to counteract useless complexity gain of the programs, but the average time for learning the kernels will necessarily increase.

\subsection{Transferability of the evolved kernels} \label{sec:transfer_results}

An important question to analyze is whether the kernels evolved by MOECov are only valid for the sentiment datasets in which they have been learned or they can also be used to predict sentiment in the other datasets. This question can be frame on the general research that investigate the transferability of solutions found by evolutionary algorithms \cite{iqbal_improving_2016,garciarena_evolved_2018,santana_structural_2012}. In order to answer this question, we have used the best programs learned for the \texttt{anger} dataset to make predictions in the other datasets. This can be considered as a transfer learning scenario in which the \texttt{anger} dataset is the source domain and all the other datasets serve as target domains. Notice, that in this particular example we do not recompute the hyperparameter values for the kernels. We simply apply the kernels  \emph{as they are} to the target datasets. 

\begin{figure*}[htbp]
\begin{center}
\includegraphics[width=1.0\textwidth]{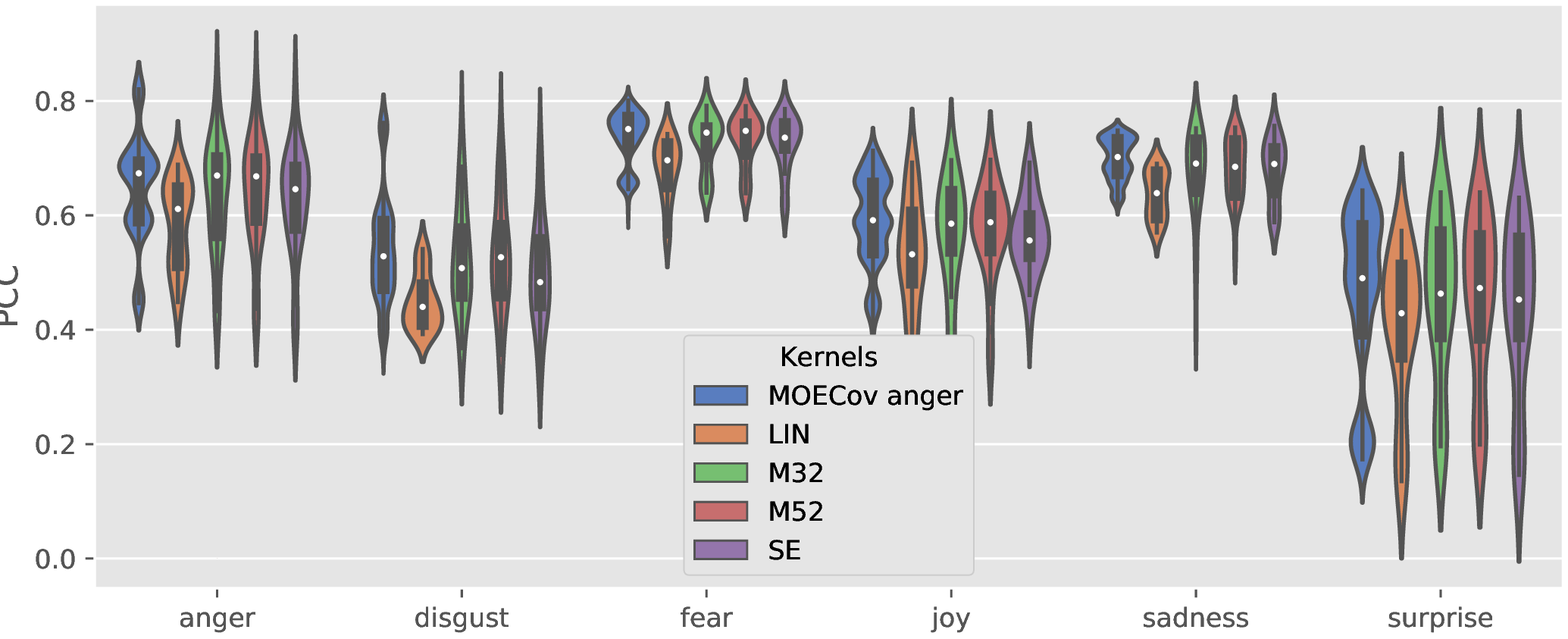}
\caption{Results of the transferability experiments for the PCC metric. Each coloured shape shows a kernel density estimation of the distribution. A boxplot of the results is shown inside. More is better.}
\label{fig:PCC}
\end{center}
\end{figure*}

\begin{figure*}[htbp]
\begin{center}
\includegraphics[width=1.0\textwidth]{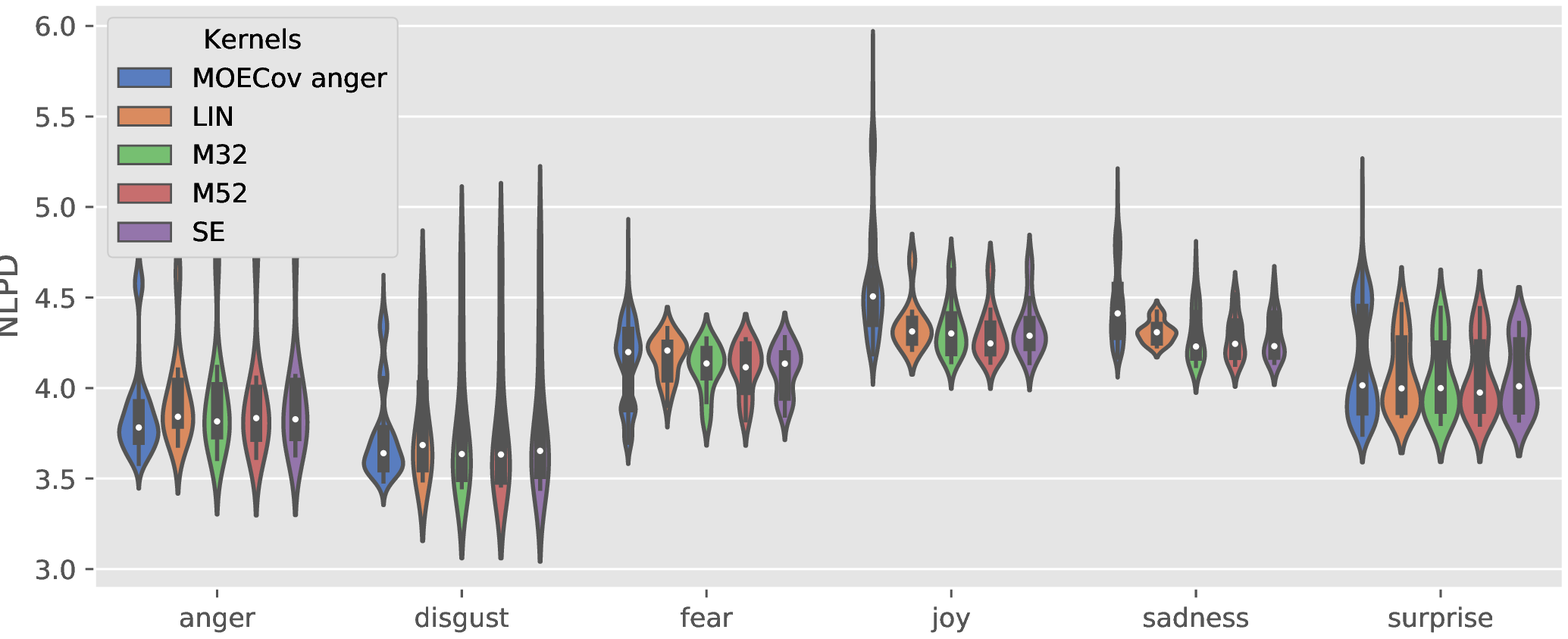}
\caption{Results of the transferability experiments for the NLPD metric. Similar to the previous figure, a boxplot is shown inside each coloured shape that shows the estimation of the distribution. Less is better.}
\label{fig:NLPD}
\end{center}
\end{figure*}

Figure~\ref{fig:PCC} and Figure~\ref{fig:NLPD} respectively show the distributions of the objectives values obtained of this  experiment for the PCC and NLPD metrics. In the figures, MOECov\_anger indicates the kernels learned using  the \texttt{anger} dataset. Notice that all the other algorithms have been learned using (training) data for each target dataset.  The analysis of the figures indicate that the transferability of the kernels depends on the type of metric used. Results for PCC are at least as good as those obtained with the other kernels. However, for the NLPD metric results are slightly worse. 
  
The main conclusion from this experiments is that the kernels evolved for predicting some sentiment can be also useful to predict other  sentiments. This means that the embeddings contain the relevant information for the prediction and that the type of transformations that make a kernel a good predictor are similar across sentiment domains.

\section{Conclusions} \label{sec:CONCLU}

Sentiment classification is a relevant problem in NLP. In previous work it has been shown that GaussProc regression is an efficient approach to solve this problem when considering predicting the sentiment with a finer level of detail, as a regression problem. However, our hypothesis was that the fixed structure used by classical kernels lacks the flexibility to capture more subtle differences in datasets. Therefore, in this paper we have proposed MOECov as way to evolve the structure of the kernels. By addressing the creation of kernels as a multi-objective problem we have been able to generate kernels that simultaneously optimize two of the accuracy metrics previously proposed, and in addition are optimized for computational complexity.

As far as the authors are concerned this is the first work that uses evolved GaussProcs for NLP problems. We had not found either previous studies that tries to optimize the hyper-parameters of fixed kernels simultaneously considering two or more metrics. 

There are a number of ways in which our studied could be extended.  Other sentiment datasets, possibly in other languages, could be considered. Furthermore, other semantic classification tasks, such as, the post-editing effort prediction could be addressed using our approach. We also noticed that at the time of stopping the MOECov algorithm the quality of the solutions was still improving. Therefore, more fitness evaluations are likely to produce better results of the algorithm. Furthermore, automatic tuning of the algorithm's parameters should be accomplished since we did not tune the parameters of the algorithm for the present study.

\section*{Acknowledgments}
The research presented in this paper is conducted as part of the project EMPATHIC that has received funding from the European Union's Horizon 2020 research and innovation programme under grant agreement No 769872 \includegraphics[width=0.03\columnwidth]{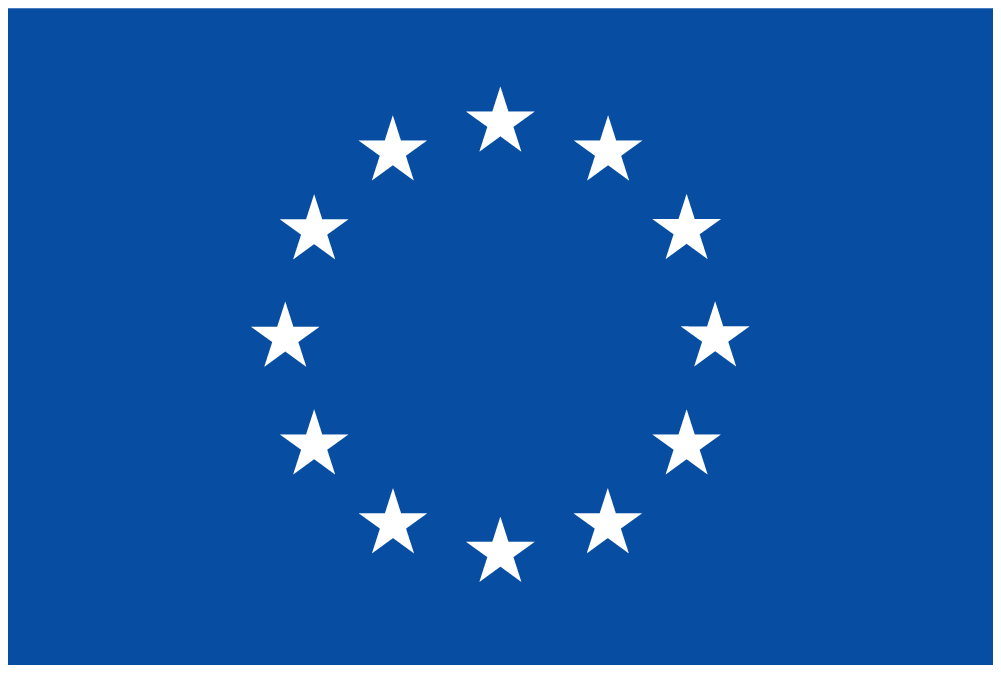}. It has also been partially supported by the Basque Government (ELKARTEK programs), and Spanish Ministry of Economy and Competitiveness MINECO (project TIN2016-78365-R). Jose A. Lozano is also supported by BERC 2018-2021 (Basque Government), and Severo Ochoa Program SEV-2017-0718 (Spanish Ministry of Economy, Industry and Competitiveness).

\bibliographystyle{unsrt}  
\bibliography{references}

\end{document}